# Ethical Implementation of Artificial Intelligence to Select Embryos in *In Vitro Fertilization*

Michael Anis Mihdi Afnan
Department of Medicine
Imperial College London
London, UK
michaelafnan@icloud.com

Cynthia Rudin
Departments of Computer Science,
Electrical Engineering and
Statistical Science
Duke University
Durham, North Carolina, USA
cynthia@cs.duke.edu

Vincent Conitzer
Departments of Computer Science,
Economics and Philosophy & Institute
for Ethics in AI and Departments of
Computer Science and Philosophy
Duke University & Oxford University
Durham, North Carolina, USA
conitzer@cs.duke.edu

Julian Savulescu
Uehiro Centre for Practical Ethics & Wellcome
Centre for Ethics and Humanities & Murdoch Children's
Research Institute
Oxford University & Oxford University & Royal Children's
Hospital
Oxford, UK
julian.savulescu@philosophy.ox.ac.uk

Abhishek Mishra
Uehiro Centre for Practical Ethics
Oxford University
Oxford, UK
abhishek.mishra@philosophy.ox.ac.uk

Yanhe Liu
Monash IVF Group & School of Human Sciences
& School of Medical and Health Sciences
Monash IVF Group & University of Western Australia
& Edith Cowan University
Southport, Australia
gift0409@yahoo.com.au

Masoud Afnan
Department of Obstetrics and Gynaecology
Qingdao United Family Hospital
Qingdao, China
masoudafnan@me.com

**ABSTRACT**

AI has the potential to revolutionize many areas of healthcare. Radiology, dermatology, and ophthalmology are some of the areas most likely to be impacted in the near future, and they have received significant attention from the broader research community. But AI techniques are now also starting to be used in *in vitro fertilization (IVF)*, in particular for selecting which embryos to transfer to the woman. The contribution of AI to IVF is potentially significant, but must be done carefully and transparently, as the ethical issues are significant, in part because this field involves creating new people.

We first give a brief introduction to IVF and review the use of AI for embryo selection. We discuss concerns with the interpretation of the reported results from scientific and practical perspectives. We then consider the broader ethical issues involved. We discuss in detail the problems that result from the use of black-box methods in this context and advocate strongly for the use of *interpretable* models. Importantly, there have been no published trials of clinical effectiveness, a problem in both the AI and IVF communities, and we therefore argue that clinical implementation at this point would be premature. Finally, we discuss ways for the broader AI community to become involved to ensure scientifically sound and ethically responsible development of AI in IVF.

## 1 Introduction

*In vitro fertilization* (IVF) is a clinical technique which has revolutionized the treatment of infertility. The process involves fertilizing the egg in a laboratory and replacing the resultant embryo into the uterus. Natural fertilization and conception is an inefficient process, with low chances of a live birth for any particular embryo. The solution both in nature and with medical treatment is to create multiple embryos, so that ultimately one will probably implant. In nature, the cost is time to pregnancy or, in the event of no embryos implanting, the pain of childlessness. In clinical practice, the cost is additionally measured in dollars. To increase the efficiency of clinical practice, much attention has been given to selecting the embryo that is most likely to implant. A recent innovation in the laboratory is time-lapse imaging of the embryo in culture over a number of days. This gives rise to thousands of visual data points, and with it the promise of augmenting the embryo selection process with artificial intelligence (AI)-based models. In this paper, we provide an overview of the IVF process, review current approaches to using AI in embryo selection, discuss ethical issues of using AI in this specific field, and make proposals for the ethical implementation of this new technology. We finish with encouragement for AI researchers to collaborate with fertility clinicians to take this research forward in a meaningful and ethical way.

## 2   The In Vitro Fertilization (IVF) Process

Each year, millions of couples who suffer from infertility pin their hopes of starting or growing their family on IVF [1]. Heavily criticized by many at first as an unethical human experiment [2], the technique has become one of the most successful therapeutic innovations of the past half-century, leading to over 9 million babies born since the first IVF birth in 1978 [1]. The limit of IVF's success, however, is reflected in the millions more whose hopes have not been fulfilled. Particularly for those with advancing age and comorbidities, but also for every couple who tries, success is not guaranteed. On average, across all age groups, the live birth rate per treatment cycle is 26.1% in the UK [3].

To maximize the chance of retrieving a good quality egg and subsequent embryo, women are given hormone treatment to stimulate development of multiple eggs, which are then harvested, inseminated and the resultant embryos cultured in the laboratory for 2-6 days. Typically 2-4 would reach the blastocyst stage around day 5 or 6. The embryologist would then select 1 blastocyst for transfer to the uterus. Any unused embryos thought to be viable are then frozen for use later in case the initial transfer following egg collection is unsuccessful, or if successful, for a future sibling [4]. It is important to note that competitive embryo selection is unique to IVF, and does not occur in nature. IVF clinics are keen to improve on current embryo selection strategies in order to maximise the chance of pregnancy at an early stage of the couple's treatment.

Early embryo development at the preimplantation stage is a very dynamic process. Hours after fertilisation, two pronuclei are formed carrying DNA material contributed by the sperm and the egg. The pronuclear membrane breaks down shortly before the first cell division, leading to a 2-cell embryo. As cells continue to divide, they become more compact with increased cell-to-cell interaction from 3 days post-fertilization. On day 4, the embryo reaches the "morula" stage, where borders between cells become invisible. During the next 1 to 2 days, cells separate into 2 layers, with a growing cavity formed between them; at this point, the embryo is called a "blastocyst". The outer layer of cells, also known as the trophectoderm, will become part of the placenta following implantation, while the inner layer (inner cell mass) will become the fetus. Both layers hatch out of the shell around the embryo (zona pellucida) before implantation into the uterus.

Traditional embryo selection is based on several snapshot observations of an embryo under a microscope, at specific time points during culture (Figure 1). Considering the dynamic nature of embryo development, the static nature of the information collected in this method limits the accuracy of embryo selection [5].

Examples of other developments to select embryos more likely to implant include (1) allowing embryos to self-deselect via extended culture in the laboratory [6], (2) metabolomic profiling of spent culture media [7] or (3) adding extra genetic testing such as pre-implantation genetic testing for aneuploidy (PGT-A) which is controversial because of its invasive nature and diagnostic imperfection [8].

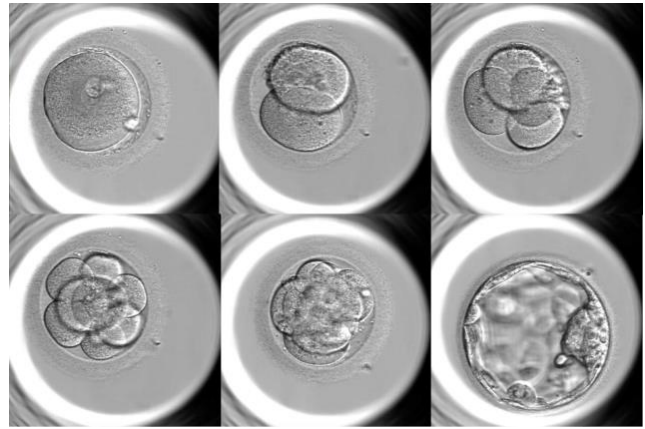

**Figure 1: Embryo development from a few hours after fertilisation (top-left) with 2 pronuclei to the "blastocyst" with trophectoderm and inner cell mass (bottom-right).**

Evaluation of embryo quality by the embryologist is limited by considerable inter-operator variability, due to the current mix of objective and subjective measures in assessment, and human factors, such as being influenced by confounders [9]. However, the availability of numerous data points made available by recent advances, such as with time-lapse videography [10], has enabled novel parameters to be measured for embryo selection [11, 12, 13]. Debate is still ongoing regarding the best approach of using such time-lapse images for embryo selection [14].

## 3 AI as an Embryo Selection Tool to Improve the Success Rate Per Transfer

The application of AI in IVF has the potential to provide more objective, more rapid, and potentially more accurate evaluation of key steps in the IVF process, to make it more reproducible and repeatable when compared with a purely human approach [15]. In particular AI for embryo selection has attracted much interest, and potentially holds much promise [16].

The type of AI that can help embryo selection is Machine Learning (ML) – models that can automatically learn and adapt as they are exposed to more data (whether images or other data). This is particularly useful when there is access to lots of data, but we do not immediately know how to leverage it to make better predictions, or when we cannot manually process it all to generate meaningful knowledge. Potential variables include morphological features, such as cleavage of the embryo cells (blastomeres), fragmentation, morphokinetic features (including time intervals between certain features), and clinical factors such as age of the woman or cause of infertility [17]. Computer Vision (CV) allows large amounts of image data to be automatically analyzed by algorithms, and rapid recent advances in this field offer great promise to improve embryo selection.

## 4 Current State of Research on the Use of AI to Select Embryos

We searched MEDLINE, Embase and Google Scholar for full-text studies evaluating AI to select embryos using the strategy included in the appendix. We checked the citations of papers we identified in the search for any publications we might have missed.

Studies evaluating AI for embryo selection make impressive accuracy claims for their ML models [18, 19]. One commonly reported performance measure is the receiver operating characteristic (ROC) curve which shows how a test's sensitivity and specificity correlate at different thresholds. The area under this curve (AUC) indicates the test's performance. An AUC >0.9 usually indicates outstanding performance, and the ML models from the studies cited above surpass this benchmark.

Studies that evaluate the efficacy of AI models for embryo selection do so for 2 types of outcomes; a) outcomes meaningful to the patient, such as a live birth or a fetal heartbeat positive pregnancy, or b) agreement with the existing standard, which in this case would be assessment by embryologists. One of the challenges of using live birth as the meaningful outcome (ground truth) is that a potentially viable embryo can result in either a live birth, or no live birth, depending on other, non-embryo factors, such as the health of the mother.

Tran, Cooke and Illingworth's [18] study belongs to the former category. They evaluated a model called IVY which rates how likely an embryo is to lead to a fetal heartbeat (FH) pregnancy on a confidence scale of 0 (definitely will not implant) to 1 (definitely will implant). Their ROC curve's AUC was 0.93. However, as Kan-Tor, Ben-Meir and Buxboim [20] point out, the majority of the embryos on which the algorithm had been trained and tested were of such poor quality that they would have been discarded in any event, thereby artificially inflating the AUC. As Kan-Tor, Ben-Meir and Buxboim explain, the clinical need is to identify the embryo with the highest chance of success among a set of embryos that appear to be potentially viable, and not from embryos which embryologists readily discard.

Khosravi et al.'s study [19], on the other hand, belongs to the latter category. They categorized embryos into 3 groups – good-, fair-, and poor-quality embryos according to a consensus of multiple embryologists. They then evaluated their AI algorithm's ability to identify the good- and the poor-quality embryos (but not the fair-quality embryos); for this task the algorithm achieved 96.94% accuracy. This was better than the performance of individual embryologists. However, broad categorizations into "good" or "poor" quality are of limited benefit when trying to find the best embryo in a group of similar-quality embryos.

The above analyses of Khosravi et al. [19] and Tran, Cooke and Illingworth's [18] studies demonstrate the importance of understanding exactly how researchers test their algorithms before drawing conclusions from headline statistics. These studies are important steps to investigate efficacy (the ability to produce a specified outcome in experimental circumstances), to develop the tool, and establish proof of principle. However, they are only a prelude to testing in the clinic. When Curchoe et al. [21] reviewed how the results of AI studies in reproductive medicine relate to real-life clinical practice, they highlighted four pitfalls that are common throughout the literature: small sample sizes, imbalanced datasets, non-generalizable settings and limited performance metrics.

Furthermore, to date, no AI studies for embryo selection using a Randomized Controlled Trial (RCT) have been published, though 1 is registered [22]. The lack of RCTs appears to be typical of much of AI in medicine [23]. The problem of lack of evidence before implementation is compounded by the IVF industry which is notorious for aggressively marketing unproven clinical and laboratory "add-ons" [24, 25]. The problem is compounded because clinicians who do not have an adequate understanding of AI will find it difficult to critically navigate the literature which contains unfamiliar concepts and terminology.

Many AI studies for embryo selection use uninterpretable ("black-box") machine learning models. These models are either too complicated for any human to understand, or they are proprietary – in which case, comprehension of such a model is not possible for outsiders [26]. Specifically, many studies in this field use neural networks that are not interpretable, and not designed to be interpretable (e.g., Chen et al. [27]). Other approaches use interpretable features (whether they are labeled manually by doctors or labeled by neural networks whose output can be manually verified) but combine them in uninterpretable ways, such as using principal component analysis (PCA) pre-processing (which forces a dependence on all variables) followed by a machine learning method such as a neural network or random forest [28, 29]. The work of Leahy et al. [30] is interesting because the model is decomposable into separate neural network models that each extract different information (e.g., measurements of an embryo) that can be directly checked by an embryologist. Combining these separate models into an interpretable "combined" model to form the final prediction would be something we could potentially recommend; if their final combined model was interpretable (such as a scoring system or sparse decision tree), then each piece of the whole system would be directly checkable by an embryologist for correctness, and thus interpretable. A third category of studies use fully interpretable features (e.g., measurements of the embryo taken by embryologists), but use older techniques that are not particularly accurate and do not explicitly optimize for interpretability (for instance, the models are not sparse). These works generally do not apply any computer vision techniques, relying instead on humans to estimate measurements from the embryo images. Examples include the works of Raef, Maleki and Ferdousi [31] and Morales et al. [32], who created interpretable hand-calculated features and applied a variety of classical machine learning algorithms to them.

The opaqueness or 'black-box' nature of AI models is problematic for two main types of reasons: ethical, and epistemic, which we will describe next.

## 5 Ethical Concerns with Black-Box AI Models

### 5.1 Failure to Perform Randomized Controlled Trials

The most important ethical issue facing the adoption of AI assisted IVF is the need for careful RCTs against best current approaches. Whilst 1 RCT has been registered [22], it is premature to implement a technology in the clinical setting before the trial results are made available. No matter how promising a new intervention appears to be, the gold standard for evaluation is the RCT. Failing to do such trials risks harming patients, as does failing to perform systematic reviews of existing evidence and failing to publish negative results [33].

Both black-box and interpretable models must be systematically studied using RCTs. Even highly promising interventions can do more harm than good [33].

### 5.2 Equipoise

The ethical justification for RCTs is that *equipoise* exists between the proposed new intervention and existing care. Equipoise exists when the new intervention is not known or reasonably believed to be significantly better or worse than the existing intervention. For embryo selection, equipoise would therefore exist if the use of AI for selection is neither known to be better or worse in achieving successful pregnancies than existing approaches. From the literature, it is very difficult to determine whether equipoise exists for AI models, and black-box models in particular: one would need to determine exactly when AI would be used (e.g., for obvious cases? For non-obvious cases only?), how it would be used in the context of the clinic (e.g., does the clinician always follow the AI prediction?), and for what populations and what settings the reported AUCs would be representative. As we discuss below, black-box models have serious problems with robustness, and as discussed above, AUC values reported from published studies cannot necessarily be trusted. Also, since interpretable AI methods have not been heavily developed, we cannot assess the value of interpretability in the decision-making process. In other words, we currently cannot assess whether a human/AI "centaur" could be better than either one alone. This information would be essential to the question of equipoise and whether to conduct an RCT.

Validation studies would help to assess these questions; typically these are conducted on a new sample from a different clinic than the AI method was trained on. However, we are not aware of any such validation studies.

### 5.3 Compromised Shared Decision-Making

An important concern centers on the use of opaque AI models potentially compromising shared decision-making and patient-centered medicine more broadly. Over the past few decades, the accepted model of clinical practice has shifted from a paternalistic one, where the clinician's opinion and recommendation are simply accepted by the patient, to one of shared decision-making where this power and informational asymmetry is reduced to the benefit of patient autonomy [34]. Clinical AI models that are opaque (so that medical explanations for a model's recommendation are inaccessible) compromise this shared decision-making due to the inability of the clinician and the patient to understand the model's decision [35]. While there have been some counterarguments raised as to whether shared decision-making is truly compromised by opaque AI models [36], application of AI in embryo selection should be guided by an awareness of such potential dangers. It will be important to fully explain what is known about how the AI model comes to a "decision" (nature and size of dataset, reasons for confidence in prediction, possible alternative lines of justification, etc.), and further examine how interactions between clinicians and patients may change, both at the point of embryo selection as well as at the point of implantation failure. Clinicians should explain the basis of how embryos are selected for transfer, whether it is clinical or AI-assisted. If information that is traditionally conveyed to the patient as to why a particular embryo is selected – for example the number and symmetry of the cells, or if the cells are fragmented, and therefore what the chances of implantation are, and why implantation may fail – are no longer accessible, shared decision-making might indeed be compromised. Existing measures of shared decision-making and decision quality, such as the Decision Conflict Scale [37], the OPTION Scale [38], and the SURE Test [39] (among other patient-reported measures) can be used to guide such an evaluation.

It is important, however, not to overstate this concern. Firstly, AI-assisted decision-making should be compared to the status quo. Current expert judgment is based on biologically meaningful measures, which, although more broadly communicable than decisions of opaque models, are not very accurate for predicting a live birth. AI-assisted decision-making may not be worse (but may also not be better). More importantly, autonomy requires understanding information relevant and meaningful to one's values. Knowing the basis of a prediction (cleavage rate, symmetry, etc) is not relevant: what is relevant are the risks, side-effects and benefits, and the confidence attached to these assessments.

However, black-box AI has the potential to significantly undermine shared decision-making in a way that interpretable AI does not. Although they are marketed and approved as decision aids to the clinician, where the decision finally rests with the clinicians, black-box AI will in practice have the potential to introduce a new form of paternalism: "machine paternalism." It is known that people tend to be complacent about use of automation [40] and tend to be accepting of AI in a role when they are familiar with AI in that role or when they believe it performs well in that role [41]. In practice it is hard to see how clinicians will challenge the deliverances of black-box AI. Indeed, doing so without good reason might open them up to legal liability. So in practice, black-box AI risks instrumentalizing the clinicians in a way that interpretable AI does not. While interpretable AI is an enhancement of human decision-making, black-box AI is a replacement for it.

### 5.4 Misrepresentation of Patient Values

Another ethical issue concerns potential harms from a misrepresentation of patient-values in the decision process. For example, there are reported differences between early morphokinetic profiles between male and female embryos [42, 43, 44, 45] (and other traits might be similarly differentially represented at this early stage). Models for embryo selection run the risk of systematically selecting for these traits if they are perceived by the model to be correlated with implantation success. For example, if a patient prefers that sex be randomly selected, this model may run counter to those values. If such models are opaque, this systematic favoring of particular traits may not be detected at the time of decision-making and so cannot be corrected for in a way that it could be with interpretable models. If we know exactly which (interpretable) characteristics of the embryo are correlated with various traits, it would be much easier to detect them at the time of decision-making. Otherwise, it may take considerable time for such a bias to be detected as a result of the use of AI because it will require large scale, statistically significant changes to manifest in evaluation, and a causal analysis that determines that the cause of bias was indeed the use of AI, and not another source.

If some of these traits are ethically salient ones for the patient, then this creates a scenario where the patient's values may not be sufficiently represented to guide the decision-making process for embryo selection. Such concerns have also been raised for other clinical models [46], calling for the design of such systems to be 'value-flexible' so that in clinical settings, both clinicians and patients are (1) aware of what metrics are driving a model's recommendations (either directly or as a proxy for some other medical fact/trait), and (2) able to appropriately reflect the patient's values in the decision process either directly through the model, or in subsequently adjusting the recommendation.

Again, it is important not to overstate this concern. The patient's own values could be inserted into AI algorithms (e.g., preference for sex and other non-disease characteristics) and AI might bring to the surface the importance of these values in decision-making. Of course, valuing and selecting non-disease traits (such as sex or intelligence) raises the debate around designer babies, but some have argued that such selection is permissible [47] or even a moral obligation when it relates to the well-being of a future child [48, 49, 50]

### 5.5 Health and Well-Being of Future Children

Such potential biasing of AI-selection might also have impacts on the health or well-being of future children. For example, it is possible that some disadvantageous trait (such as increased risk of cancer or mental disorder) correlates with a higher chance of implantation. However, this risk might be present unknowingly in ordinary clinical decision-making. This also underscores the importance of clinical trials not merely measuring implantation or even healthy live birth but long-term well-being of the child created by IVF through long-term (decades) follow up.

Reproduction is also unique because selection determines who will come into existence. This creates the so-called "non-identity problem" which has spawned decades of unresolved philosophical debate, sparked by [51]. Imagine Embryo A has a higher chance of implantation but unknowingly a higher chance of cancer later in life than embryo B. AI selects A. A is born but gets cancer at the age of 30. Was A harmed by the decision to select A rather than B? No, a different person (B) would have been otherwise selected. Provided that the disadvantageous trait or genes do not make A's life so bad as to have been not worth living, then A cannot be harmed by selection. On this ground, greater risks can be taken in embryo selection than with interventions on a specific embryo (such as A) which do risk harm to a specific individual [52]. Nonetheless, some have argued that parents (and clinicians) still have a moral obligation to select the embryo with the best chance of the best life [48, 49, 50].

## 5.6 Impacts of Disvaluing Disability

There is a general problem with embryo selection raised by disability activists: any kind of selection based on predicted health or well-being discriminates against the disabled and expresses a negative message about the value of their lives - the expressivist objection [53]. For example, screening for Down Syndrome has been said to express a negative view about the value of people with Down Syndrome [54]. This applies not only to AI selection but to clinical selection more generally, and there are numerous responses [53]. However, AI might considerably expand the scope of this objection: any trait that lowers chance of implantation might result in selection against that group, e.g., sex as previously discussed.

The best response to these concerns would be to monitor such effects and ensure social responses that reinforce the equality of all people, including people with disabilities. Thus, rather than forgoing selection, it is better to ensure there are sufficient social resources so that all existing people have a reasonable chance of a good life [50].

Interpretable AI may allow issues of bias to be identified earlier and in a more actionable way. Specifically, if we discover interpretable features that can be linked to various outcomes, it is easier to monitor for possible bias and avoid it.

## 5.7 Societal Impacts of AI for Embryo Selection

Successful AI models might be deployed at scale, and if such models systematically favor certain traits represented in early morphokinetic profiles, this might impact society. Even if would-be parents might not care about the sex of their future child and might be willing to accept a higher likelihood of one sex for a higher likelihood of implantation success, this will still have societal ramifications through a skewed population ratio. The scale of these ramifications will correlate with rates of IVF use in the future; the more individuals opt for IVF, the greater the impact. While such possibilities are at this stage mostly speculative, they represent a scale of impact that is significant and should therefore be considered. Since black-box models do not aim to identify specific aspects of an embryo with specific traits, it might make these issues more difficult to detect until it is too late and there are societal-level impacts. In this regard, interpretable AI may allow earlier detection of systematic favouring of certain traits.

Further, if AI-assisted IVF works better for some races than others, this could have serious societal implications. While such differences should be monitored whether a black-box or interpretable AI method is used, interpretable AI may help to detect such issues more easily, for instance, if the AI model is known to leverage factors that differ among ethnic groups (e.g., the relationship of age to fertility).

## 5.8 Black-Box Models Pose a Responsibility Gap

The final ethical issue concerns a potential erosion of ethical and legal accountability through the use of opaque AI models. If it is determined that clinicians cannot be held responsible for injuries sustained by the patient due to a reliance on opaque AI models, the responsibility for this class of errors would need to be borne by another agent. In the absence of institutionalized accountability mechanisms that hold other agents, like model developers, responsible, this creates a 'responsibility gap' when it comes to the use of AI models.

The most straightforward case in which accountability is required would be repeated implantation failure or low success rates due to suboptimal embryo selection processes, and/or injury being sustained by the patient as a result of implementation of a model recommendation (either to the mother through surgical complications or the child when he/she is born - wrongful life or birth). Under such circumstances, if the patient seeks an account of what happened or advances a charge of negligence against the clinician, the decision-making process needs to be explicable. Traditionally, if a charge of negligence is advanced, experts assess the clinician's decision-making process, and depending on whether they deem this to be medically reasonable, the clinician is either acquitted or held culpable. If AI models used for embryo selection reason in uninterpretable ways, it is unclear how a court might evaluate the doctor's decision-making, and subsequently, it would be unclear how responsibility for injury may be adjudicated [55, 56].

## 6 Epistemic Concerns with Black-Box AI Models

There are technical challenges posed by black-box or opaque systems; it is unclear how we might assess the reliability of the model's predictions, eliminate potentially confounding factors at the decision-making point, and assess to what extent the model's accuracy is representative in a given use-case. In a field where 'add-on' clinical offerings are already widespread despite inadequate evidence of effectiveness, this epistemic problem is especially troubling [25].

### 6.1 Black-Box Models Create Information Asymmetries

The use of black-box models creates an information asymmetry between the company selling the tool and the clinicians having to make daily decisions as to which embryo to transfer. Using such models would force the embryologist to abrogate decision-making to programs they themselves do not understand. It is not possible to fully evaluate whether to trust these complex models without an understanding of their reasoning processes.

## 6.2 Confounders are Rampant

If we do not understand what a black-box model is doing, it is entirely possible that its predictions are based on confounders that should not be used as predictors. Confounders are often difficult to detect and cause models not to generalize. When coupled with a poor choice for evaluation metric, the confounding might not be noticed [57, 58].

Let us construct a simple example where an obvious confounder and a standard (but ill-chosen) evaluation metric provide a situation where a useless model would appear to be excellent. In this example, the confounder is the mother's age, and the metric is overall AUC (not the AUC for an individual couple). It is largely possible that the mother's age is a major factor in predictions; what if it were the sole factor, so that a model based on an image of the embryo is predictive of mother's age only? If the model were a mere proxy for age (and we did not know it), it would be entirely useless in discriminating between embryos from the same couple, yet it may still score highly on AUC because age alone is predictive of success in implantation, and we may be fooled into thinking that it is a good model. Here there are two problems that combine to be worse than either alone: a mismatched evaluation metric, and an inscrutable model that does not reveal the problem with either the predictions or the metric.

## 6.3 Real-Time Error-Checking is Harder with Black-Box Models

The two problems discussed above (information asymmetry and the possibility of confounders) lead to a third problem, namely the difficulty of error-checking the model in real-time as it makes predictions in the clinic. We would want the clinician to be able to determine whether the model is reasoning in a way that is obviously wrong and catch new problems immediately should they arise. For instance, after a change in camera setting, an algorithm might suddenly start thinking that the shape of a current embryo looks like an embryo from the training set with a completely different shape. A clinician could potentially catch that problem immediately if they knew the reasoning process of the model.

## 6.4 The Economics of "Buying Into" a Brittle Model Does Not Favor Clinicians or Patients

A potential consequence of the problems of information asymmetry and confounders listed above would be that black-box model performance may be brittle to changes from the system it was trained on, and thus would likely be limited to the ecosystem in which it has been shown to work. This means that a clinic using this model would need to buy into that ecosystem, ovarian stimulation regimens, use of incubators and culture medium amongst other potential variables. This gives AI companies a great deal of *economic power* over clinics, potentially increasing treatment costs.

## 6.5 Overall Troubleshooting is Difficult for Black-Box Models

If the model were more interpretable, it might be easier to troubleshoot broad problems in the model (beyond serious issues that might be noticed in real-time usage). This includes ethical concerns such as sex, disability or racial bias (as we discussed), as well as epistemic issues with accuracy or subtle confounding. If interpretability reveals flawed reasoning processes, the designer would be forced to alter the model to use correct reasoning, leading potentially to more robustness across ecosystems.

## 7 Interpretable ML as the Way Forward in Embryo Selection

An interpretable ML model is a predictive model that is constrained so that a human can better understand its reasoning process [26, 59]. Interpretable ML is a field that dates to the beginning of AI, back to the days of expert systems. The benefits of interpretable models are clear: by understanding the reasoning processes of predictive models, physicians can troubleshoot them and justify their decisions (to patients, other physicians, and during lawsuits). They would not need to place blind trust in a black-box model. Physicians and interpretable ML models can create a "centaur" that leverages both the information in a database (through ML) and a human's system-level way of thinking about problems. Furthermore, correlating patient values (chance of disability, sex, single vs. double embryo transfer and chance of implantation) with outcomes can be more easily accommodated by interpretable models.

Focusing on increasing the use of interpretable AI models is an elegant approach to both the epistemic and ethical concerns, by dispelling the opaqueness and allowing precise explanations of model predictions. For instance, models that are not opaque have an advantage because their use preserves existing mechanisms of accountability to a greater extent by allowing clinicians to understand model decisions better and thus retain the responsibility. For models that are opaque, revisions to such mechanisms would be necessary. This argument is far from resolved, but it is a promising reason to favor interpretable models over black-box ones.

To the extent that there exists little or no trade-off between how interpretable a model is and how accurate it is for embryo selection, interpretable models are thus a promising solution. If it is the case that there is a salient difference in performance between interpretable and non-interpretable models, alternative solutions to both of the above epistemic and ethical concerns might have to be developed, so that we may benefit from the higher predictive accuracy of non-interpretable models. For now, there is no reason to believe that a salient performance difference between interpretable and non-interpretable models would exist. Interpretable models perform just as well even for benchmark datasets in computer vision, and we will go into more detail on this important point shortly. In fact, interpretable models are easier to troubleshoot (as domains change, as unusual cases arise, as cases indicating bias need to be investigated), and thus lead to overall better performance of the model.

A major question in interpretable ML is what interpretability metric to use, as these metrics must (by definition of interpretability in that domain) be domain dependent. For computer vision for natural images, there have been major successful efforts by numerous groups of researchers to create interpretable neural networks that do not lose accuracy over their black-box counterparts. These neural models go well beyond modeling only the "attention" of the network (that is, where the network is looking within an image), and are particularly useful for computer vision problems. Such interpretable neural networks could use different types of logical reasoning processes, including:

• Case-based reasoning (variations of k-nearest neighbors): In this case, the network would point out which parts of a test image are similar to prototypical past cases. The prototypical cases are chosen by the network along with the ways in which images are similar to each other (e.g., Chen et al. [27]). One could envision an embryologist looking at a test image of an embryo, with an interpretable ML model pointing out how parts of it look similar to other prototypical known embryos whose outcome is known. Case-based reasoning models have been developed for different problems in radiology using interpretable ML [60, 61], and thus such methods would be poised for use in embryology.

• Latent space disentanglement, where all information about a single concept (such as mother's age, or embryo size, density or color) is forced to travel through a single node of a network. Another way to say this is that each axis of a latent space (where an axis corresponds to activation of a node) represents a concept. This helps to understand information flow through the network (e.g., Chen, Bei and Rudin [62]). These types of disentangled models could potentially be useful for separating out the type of equipment, the age of the mother, and other pieces of information that might be embedded within the image of the embryo.

• Networks imbued with logical structure, e.g., probabilistic decision trees. By forcing the network to reason logically, we may be better able to understand its reasoning process (e.g., Wu and Song [63], Li, Song and Wu [64]).

There are many challenges still in designing interpretable neural networks, particularly when the domain experts themselves do not know what constitutes interpretability; in other words, there are many directions for future research.

For problems involving categorical or real data ("tabular" data, rather than image data, time sequence data, or text data), interpretable ML models can also be developed. These can potentially take the form of a medical scoring system, which means a small number of integer "point" values that sum, and translate into a risk (e.g., Ustun and Rudin [65]). For tabular data, neural networks and other forms of black-box models do not seem to provide additional accuracy, which means that optimized medical scoring systems might be as accurate as one could get (depending on the dataset) [66, 67].

An interesting direction for future research is to combine interpretable neural networks for computer vision (to handle the visual data, similar to the work of Leahy et al. [30]) with interpretable models for tabular data (rule lists or decision trees, for instance) to form a global interpretable model that handles these heterogeneous data types.

One key point that has emerged from past research is that as long as one can design the interpretability metric carefully to match the domain, interpretable models tend not to lose accuracy relative to their black-box counterparts [26, 27]. As far as we know, modern interpretable ML methods have not yet been fully applied to IVF.

In Table 1, we summarize the advantages of interpretable AI models over black-box models.

| Normative Considerations | Black-Box ML System | Interpretable ML System |
|---|---|---|
| Clinical decision-making | Replaced (machine paternalism) | Enhanced |
| "Centaur" arm of RCT which combines ML and human expertise | Not yet possible – acceptance of black-box recommendation is all-or-nothing | Permits such an arm |
| Responsibility for treatment success | Unclear | Clearly remains with the clinician |
| Biases that misrepresent patient values, and have unintended consequences for future people and society | May go unnoticed until societal effects become significant | Easier to detect earlier: the model reports the parameters on which it is basing its decision |
| Confounders which reduce the true predictive power of the model | May go unnoticed until clinics report underwhelming success rates | Easier to detect earlier: the model reports the parameters on which it is basing its decision |
| Error-checking | May go unnoticed until clinics report underwhelming success rates | Clearly-erroneous decision-making could be identified by an embryologist |
| Economics | Clinics may need to purchase specific equipment to guarantee the success of models which require the same equipment used for evaluation | Clinicians could modulate their interpretation of a model's suggestion under different conditions, if the model can explain the basis of its decisions |
| Accuracy/capability | There is no evidence to suggest that black-box models are any more accurate than interpretable models | |

**Table 1: Summary box of reasons why interpretability gains an advantage over black-box models**

## 8 Recommendations

### 8.1 Rigorous Evaluation with RCTs

Researchers must evaluate AI models to select embryos using the gold standard of RCTs against best clinical judgement or black-box AI, if these have been deployed into practice or show promising results. Key outcomes for evaluation include time to live birth, number of embryo transfers before live birth and associated cost analysis, as well as live birth per egg collection, and health of the baby. Researchers should monitor the effects of the new technology with post-implementation surveillance. Before an RCT is performed, a validation study should be performed on data from different clinics than those used to train the model.

### 8.2 Interpretable AI

Programmers should aim to build interpretable machine learning models where biologically meaningful parameters guide embryo assessment, reducing the risk of hidden biases in algorithms causing unintended harms to society, permitting better troubleshooting, and better enabling clinicians to counsel their patients on the thinking underlying their treatment.

### 8.3 Regulatory Oversight for Interpretable AI

The importance of interpretability should be captured in mechanisms of regulatory oversight. Current regulatory approaches attempt to capture medical AI models as a type of medical device; they should further require either that AI model developers not produce black-box models if interpretable models are shown to have similar performance, or that any black-box model must come with the next-best interpretable model considered and trialed. Further, despite the fact that the field of assisted reproductive technology utilizes 'good practice' regulation for many advancements (such that violations are not legally punished), this would not suit the many risks of AI in embryo selection as outlined above. A 'hard' regulatory stance that promotes interpretable models would be a more advisable approach.

### 8.4 Access to Code and Data

Data and code used to create ML models should be made publicly accessible. This would enable reproducible research and the advancement of an exciting and important academic field. A high-quality public model would, at the very least, provide a performance baseline for other models.

### 8.5 Respect for Patient Privacy and Autonomy

Procedures should be put in place for securing patient privacy when data is shared, such as data anonymization. All patients who use AI to select embryos should give fully informed consent, including knowledge of limitations and unknowns, use of data and images, and harms and benefits as shown by RCTs. They should be informed of how a model arrives at a recommendation. Where possible, patient values should be inserted into the reasoning process of selection models.

### 8.6 Involving the Broader AI Community

Many young ML researchers are eager to get their hands on data to try out the latest techniques, and are passionate about using the technology to make the world a better place. Their participation should be encouraged. Currently, datasets for embryo selection are not broadly available. A naïve release of such data may do more harm than good, potentially inviting simplistic evaluations of ML techniques that fall prey to many of the criticisms we have discussed. Releasing a dataset and suggesting evaluation criteria for it which reflects actual practice, and takes ethical concerns into account, will require a broader discussion between embryologists, ethicists, and researchers in AI and statistics, and will also require addressing privacy concerns. This discussion ought to continue after the data are released. Nonetheless, allowing the broader AI community to see the data and get involved in their analysis will ensure that flawed and biased evaluations do not easily fly under the radar. It will also likely bring other important issues into the open that we have not yet recognized.

We summarize our recommendations in Table 2 below.

---

• Use of replicable, interpretable machine learning tools and data
• Well designed and conducted RCTs
• Post implementation surveillance
• Regulatory oversight requiring interpretable AI whenever possible
• Funding for public institutions to transparently develop and evaluate machine learning models, and open access to code used in models
• Procedures for maintaining security of patient/embryo data whilst permitting ethical data sharing
• Fully informed consent to use AI
• Inclusion of patient values into AI programmes where possible
• Participation from the broader AI community

---

Table 2: Summary box of recommendations

## 9 Conclusion

Starting or growing a family is an immensely significant decision; technology which could help individuals who make that decision realize their goal would be invaluable. We see potential for AI in IVF to help couples have children earlier in their treatment and at a lower cost. However, researchers, companies and clinics must ensure that the technology they promote or adopt brings real, measurable benefits to patients and, most importantly, does no harm. In this article, we highlighted limitations of current ML models and the studies which evaluate them, we drew specific attention to the ethical concerns that this technology could introduce in its current form, and suggested a path forward in terms of model design and evaluation. Most importantly, we hope to see interpretable machine learning models that clinicians could understand, troubleshoot and explain to their patients, rigorously evaluated with RCTs. We believe these are essential for creating tools which are fit for use for real individuals, hoping to start or grow a family, in the clinical setting.

## 10 Conflict of Interest

The authors declare no conflict of interest.

### ACKNOWLEDGMENTS
The authors received no funding for this review.